\documentclass{article}
\title{Probabilistic prototype models for attributed graphs
} 
\author{S. Deepak Srinivasan, Klaus Obermayer \\ {\footnotesize Neural Information Processing Group, Technische Universitaet Berlin} \\  {\footnotesize Sekretariat FR 2-1, Franklinstr. 28/29, 10587 Berlin} \\ {\footnotesize\{deepak,oby\}@cs.tu-berlin.de}}
\date{}
\usepackage{graphicx,type1cm,eso-pic,color,verbatim}

\usepackage{amsfonts}
\usepackage{amsmath}
\usepackage{epsfig}
\usepackage{amsthm}
\usepackage{multirow}
\include{config}
\begin{document}

\setlength{\parskip}{10pt}

\maketitle

\begin{abstract}

This contribution proposes a new approach towards developing a class of probabilistic methods for classifying attributed graphs. The key concept is \emph{random attributed graph}, which is defined as an attributed graph whose nodes and edges are annotated by random variables. Every node/edge has two random processes associated with it- occurence probability and the probability distribution over the attribute values. These are estimated within the maximum likelihood framework. The likelihood of a random attributed graph to generate an outcome graph is used as a feature for classification. The proposed approach is fast and robust to noise.

\end{abstract}

\section{Introduction}

Attributed graphs are used to represent data as diverse as images, shapes, molecules and Protein structures. The statistical analysis of a dataset of patterns represented by graphical structures is a challenging problem and is closely related to tasks such as density estimation, mixture modelling, classification and clustering. There have been some efforts to develop probabilistic models for attributed graphs in the context of pattern recognition.  Wong et al., \cite{w85,w80} propose the concept of a random graph  which takes into account structural and contextual probabilities. An instantiation (outcome) of a random graph is an attributed graph, which enables the characterization of an ensemble of outcome graphs with a probability distribution. Sole-Ribalta et al., \cite{ss09} generalize the idea of random graphs to structure described random graphs (SDRG), with node and edge value distributions. Algorithms have been proposed where random attributed graph models are used for classification in the maximum likelihood framework. This framework has also been adopted by Seong et al., \cite{sk93} to develop an incremental clustering algorithm for attributed graphs, and by Sengupta et al., \cite{sb95} to efficiently organize large structural modelbases for quick retrieval. There are two features of such a definition that are quite noteworthy- (i) both the structural and contextual probabilities are considered, which are estimated with suitable independence assumptions and (ii) the ability to deal with a wide variety of attribute values. 

The present contribution aims to develop a theory for probabilistic modeling of attributed graphs similar to generative models for feature vectors and demonstrate its utility to classify graph patterns. We propose random graph models as prototypes for a set of graphs with continuous node and edge attribute vectors and estimate its parameters. Instead of using the random graph models to classify the patterns in terms of maximum likelihood, we use the likelihood values as features for classification by subsequent discriminative classifiers such as support vector machines.  

\section{Random attributed graph models}

\subsection{Definitions}

A random attributed graph (simply referred to as random graphs) is a graph whose nodes and edges are finite probability distributions. Each outcome of a random graph is a labeled graph along with a morphism of the labeled graph into the random graph.  The morphism specifies for each vertex (or edge) of the outcome graph which vertex (or edge) of the random graph generated it. The probability space of random graphs should be such that, the outcomes are attributed graphs with specified morphism relation and is complete. The definitions in this section follow \cite{w85} closely. 

Technically, the random graph $\mathfrak{G}=(\mathfrak{V},\mathfrak{E})$ \footnote{Elements of the random attributed graph are represented by $\mathfrak{fraktur}$ script.} is defined to be such that:

1. Each vertex $\mathfrak{v} \in \mathfrak{V}$ and edge $\mathfrak{e} \in \mathfrak{E}$ is a finite probability distribution

2. $\forall \mathfrak{e} \in \mathfrak{E}$, $p(\mathfrak{e} = \phi \vline \sigma(\mathfrak{e}) = \phi)=1$

3. The space of joint distribution of all random nodes and random edges is complete

Condition 2 ensures that an edge can occur in an outcome only if both its ends (terminal nodes, given by $\sigma(\mathfrak{e})$) occur. Completedness means that the space is indeed a (standard) probability space.
Consider the probability space of the joint distribution. This space is the probability space of attributed graphs and every outcome is an attributed graph.

Let ${G}=({V},{E})$ be an outcome graph. A morphism $\mu : {V} \rightarrow \mathfrak{V}$ and $\nu : {E} \rightarrow \mathfrak{E}$  specifies
the structural mapping between the random graph and its outcome. Thus, an outcome of a random graph is specified by the tuple $({V},{E},\gamma)$, where $\gamma = (\mu,\nu)$.
It is to be noted that the mappings $\mu$ and $\nu$ are into and the inverse mappings $\mu'$ and $\nu'$ are such that some elements could be mapped to $\phi$, i.e. 
$\mu'(\mathfrak{v})=\phi$ if no morphism exists. The probability of an outcome graph is then the probability of its joint outcome described by the following

\begin{equation}
p_{\mathfrak{G}}({G},\gamma)= prob\{({v}=\mu'(\mathfrak{v}), \forall \mathfrak{v} \in \mathfrak{V}, \alpha(v) =\alpha_i),({e}=\nu'(\mathfrak{e}), \forall \mathfrak{e} \in \mathfrak{E}, \beta(e) =\beta_i)\} 
\label{eqn1}
\end{equation}

where $\alpha(v)$ is the node attribute function that assigns attribute for every node and $\alpha_i$ denotes the particular node attribute value.
$\beta(e), \beta_i$ are the corresponding edge attribute function and values respectively.

\begin{figure}
  \centering
  \includegraphics[scale=0.4]{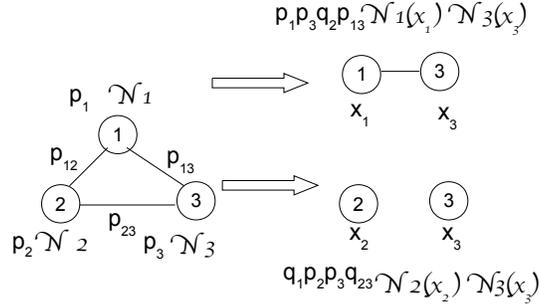}
  \caption{A random attributed graph (centre) with two outcomes and their respective likelihoods}
  \label{ex1}
  \end{figure}

We make the following assumptions to make the definition computationally feasible- node occurences are independent, and  edge occurences depend only on the nodes that the edge is incident to. 

Then, we can simplify Eq.(\ref{eqn1}) to

\begin{eqnarray}
&& p_{\mathfrak{G}}({G},\gamma)= \prod_{\mu'(\mathfrak{v}) \neq  \phi} p(\mathfrak{v}) \prod_{\mu'(\mathfrak{v}) = \phi} q(\mathfrak{v}) \prod_{\nu'(\mathfrak{e}) \neq \phi} p(\mathfrak{e}) \prod_{\nu'(\mathfrak{e}) = \phi} q(\mathfrak{e})\nonumber\\
&& \prod_{\mu'(\mathfrak{v}) \neq  \phi} prob(\alpha(\mu'(\mathfrak{v})=\alpha_v)) \prod_{\nu'(\mathfrak{e}) \neq \phi} prob(\beta(\nu'(\mathfrak{e})=\beta_e))
\label{eqn2}
\end{eqnarray}

where $p(\mathfrak{v})$ denotes a probability that the node $\mathfrak{v}$ occurs and $q(\mathfrak{v}) = 1- p(\mathfrak{v})$ is the probability that $\mathfrak{v}$ does not occur. Similar notation has been adopted for the edges as well. We note that formula in Eq. \ref{eqn2} decomposes the probability of an attributed graph instance as the product of probability of nodes/edges of generating random graphs that occur in the outcome, "not occurence" probability of nodes/edges that are absent in the outcome, and the probability of the occuring nodes/edges to assume their respective attribute values. Figure~\ref{ex1} illustrates the above definitions with an example.




\subsection{Model based clustering of attributed graphs}

The estimation of structural parameters of a random graph given a dataset follows from maximizing the likelihood: the node and edge occurence probabilities of random graphs are set to those values which maximize the likelihood of the dataset being generated from the random graph. The cost function is

\begin{equation}
E(\{G_i\},\mathfrak{W}) = \sum_{i} \ln p ({\mathfrak{W}}(G_i))
\label{eqn3}
\end{equation}

where $\ln p({\mathfrak{W}}(G_i))$ is the likelihood that random graph $\mathfrak{W}$
generates the graph $G_i$. We now consider the case where the node and edge attributes are given by feature vectors. In order to simplify the analytical treatment, we assume the  attribute vectors to be generated by Gaussian distributions whose means and covariances are to be determined. 




Initially, we maximize the cost function with respect to the node and edge occurence probabilities $p(\mathfrak{v})$ and 
$p(\mathfrak{e})$. As the node occurences are modelled by independent Bernoulli distributions, the maximum likelihood estimate is the fraction of its occurences in the dataset

\begin{equation}
p(\mathfrak{v}) = \frac{n_{\mathfrak{v}}}{N}
\label{eqn4}
\end{equation}

where $n_{\mathfrak{v}}$ is the number of occurences of node $\mathfrak{v}$ in the sample set. Similar estimates hold good for the edges except that edge occurence probabilities are normalized by their respective node probabilities (accounting for the fact that the edges cannot occur if any of their end nodes do not occur). 

\subsection{Density estimation}

We now consider the problem of estimating means and covariances of node and edge attribute distributions. It is possible to derive gradient descent update rule for the mean and covariance matrices. The vanilla gradient descent where the means and covariances are updated in the direction of the gradient (as it is assumed to the steepest direction) is not ideal as it ignores the geometry of the underlying probability space. Therefore, we use natural gradient descent to estimate the mean and covariance online \cite{am98,honk08}.

Natural gradient descent is a modification of the gradient descent procedure which takes into account the geometry of the manifold by incorporating a corrective term given by the Riemannian metric tensor. The equations for updating the means and covariances in the direction of the natural gradient are given by

\begin{equation}
\mu_{\mathfrak{v}_{t+1}} = \mu_{{\mathfrak{v}_{t}}} + \eta G_{\mathfrak{v}}^{-1} \nabla_{\mu_{\mathfrak{v}}} ln (p)
\label{eqn5}
\end{equation}

\begin{equation}
\Sigma_{\mathfrak{v}_{t+1}} = \Sigma_{{\mathfrak{v}_{t}}} + \eta H^{-1} \nabla_{\Sigma_{\mathfrak{v}}} ln (p)
\label{eqn6}
\end{equation}

where, $G, H$ are the Riemannian metric tensors in the space of means and covariances respectively. The Riemannian metric tensor in the space of mean vectors $G$ is given by

\begin{equation}
G = \Sigma ^{-1}
\label{eqn7}
\end{equation}

Hence, the online update equation for the Gaussian means is given as below

\begin{equation}
\mu_{\mathfrak{v}_{t+1}} = \mu_{{\mathfrak{v}_{t}}} + \eta (x_{v} - \mu_{{\mathfrak{v}_{t}}})
\label{eqn8}
\end{equation}

The metric tensor in the space of covariance matrices is defined as

\begin{equation}
H_{\mathfrak{v}} = \mathbb{E} ((\nabla_{\Sigma_{\mathfrak{v}}}{\mathcal{F}_\mathfrak{v}}) (\nabla_{\Sigma_{\mathfrak{v}}}{\mathcal{F}_\mathfrak{v}})^T)
\label{eqn9}
\end{equation}

which after some simplification turns out to be

\begin{equation}
H_{\mathfrak{v}} = \frac{\Sigma_{\mathfrak{v}}^{-2}}{2}
\label{eqn10}
\end{equation}

The online covariance estimation is given by a first order update rule as

\begin{equation}
\Sigma_{\mathfrak{v}_{t+1}} = (1-\eta)\Sigma_{{\mathfrak{v}_{t}}} + \eta (x_{v}-\mu_{{\mathfrak{v}_{t}}}) (x_{v}-\mu_{{\mathfrak{v}_{t}}})^T
\label{eqn11}
\end{equation}

\section{Random graphs generating classification features}

Prototype based classification schemes are widespread in the domain of attributed graphs \cite{fr08}. The key idea is to embed the graphs into a vector space in the following manner. Given a set of graphs $\{(X_{i},y_{i})\}$, we synthesize a set of prototype graphs $\mathcal{W}: \{(W_{i},y_{i})\}$ such that every graph $X_i$ is embedded in $\mathbb{R}^k$ as

\begin{equation}
X_{i} \rightarrow (d(X_i,W_1),..,d(X_i,W_k))
\label{eqn12}
\end{equation}

where $d(X,W)$ is a dissimilarity measure between the graphs and prototypes. The choice of prototypes influences the distance measure and hence the dissimilarity space. To illustrate, when the prototype graphs are chosen to be set median or means or cluster centres, it is clear as to how the distance is calculated. However, what is a suitable distance measure when we choose random graphs as prototypes ?

The key lies in defining the Kullback-Leibler divergence between the probability density of random prototype graph $\mathfrak{W}$ and the true (hidden) probability distribution  $\mathbf{q}$ \cite{ht99,dh00}

\begin{equation}
KL (\mathbf{q} \parallel (\mathbf{p}(\mathfrak{W})) = -\int \mathbf{q} \ln \frac{\mathbf{p}(\mathfrak{W})}{\mathbf{q}}
\label{eqn13}
\end{equation}
 
The unknown probability distribution $\mathbf{q}$ is represented by $\delta(g-g_i)$, where $\delta(.)$ is the Dirac delta function at every data sample $g_i$. Seperating the ln term into $\ln (\mathbf{p}(\mathfrak{W})) - \ln (\mathbf{q})$ and noting

\begin{equation} 
\int \delta(g-g_i) \ln(\mathbf{p}(\mathfrak{W})) = \ln(\mathbf{p}(\mathfrak{W_{g_i}}))
\label{eqn14}
\end{equation}
 
which is the log-likelihood that the random graph $\mathfrak{W}$ generates the outcome $g_i$. Hence, likelihood (or more precisely its logarithm) could be used as a feature for classification naturally in the dissimilarity/distance representation framework. We also note here that a feature space embedding of graphs defined by likelihood values corresponds to the framework of Jaakkola et. al., \cite{hauss98} who propose to use kernels derived from generative models.


We thus summarize the scheme as below. Given a dataset of graphs representing patterns belonging to different classes, sythesize first random attributed graphs acting as a model/prototype for each class. The largest graph (i.e. the graph with maximum number of nodes) is initialized as prototype classwise. We then present every graph in the training set, align them with the corresponding prototype and update the node and edge occurence (structural) probabilities. The means and covariances are also updated according to the formulae in Eq. (8), (11). Once the parameters of the random prototype graphs are determined, we embed the dataset into a feature space by calculating the log-likelihood between every graph in the dataset and every element in the prototype set. We point out the following notable features of this scheme: (1) More than one prototypes could be used for every class especially for datasets with diverse graphs in the same class. However, in our analysis and experiments, we consider just one random prototype per class in view of computational complexity of graph matching; (2) The size of the prototypes are bound by the size of the largest graph in the dataset (3) The number of graph matching operations during the parameter estimation stage is $N$, the size of the training set; once the prototype random graphs are sythesized, the training set (with $N$ samples) and the test set (with $M$ samples) have to be embedded in the likelihood space. This needs another $(N+M)\times K$ graph matching operations.

\section{Experiments}

\subsection{Algorithmic details}

\emph{Matching attributed graphs}- The problem of aligning random graphs with each of the sample graph and the likelihood calculation involve attributed graph matching. We adopt again the graduated assignment algorithm \cite{gr96} with a suitable compatibility function for this purpose. This algorithm minimizes a cost function as a function of match matrix over all possible matchings by an iterative procedure which estimates match matrix at every step and normalizing it. The matching quality is influenced by node compatibilites which measure how similar the nodes are structurally and attribute-value wise. In determining the morphism between random attributed graphs and outcome graphs, the compatibility function is set to the likelihood of the node being structurally present in the outcome, thus in effect finding the morphism which is most probable.

\emph{Classification procedure}- Once the random graphs have been synthesized classwise, the dataset was embedded in to a feature space by calculating the log-likelihood of graphs beng generated by the prototype random graphs. In the feature space, various classifiers were learned on the training set and validated (by performance on the validation set or by cross-validation on the training set). The classifier exhibiting best validation performance was used to classify the test data. Extensive experimentations indicated that support vector machines with polynomial/Gaussian kernels yielded the best performance. All classification experiments were done using PyML software \cite{bh08}.

\subsection{Synthetic datasets}

We first analyzed the performance of this algorithm on sythetic datasets. We consider a dataset consisting of 200 graphs in training and test set belonging to two classes. The dataset is generated by considering distortions of two base graphs classwise at different levels viz. $5\%, 10\%, 15\%, 20\%$. Node and edge attributes are generated according to a normal distribution. The noise according to the specified distortion level is added which modifies node and edge occurences and also their respective attributes. The nodes are then randomly permuted. The dataset is then divided uniformly into training and test sets. The classification scheme described in this chapter is referred to as $RAG + LF$ (Random Attributed Graph model + Likelihood as a Feature). The standard $k-$ Nearest Neighbour algorithm ($kNN$) in the graph domain is chosen as the benchmark classifier.

\begin{figure}[!htbp]
  \centering
  \includegraphics[scale=0.5]{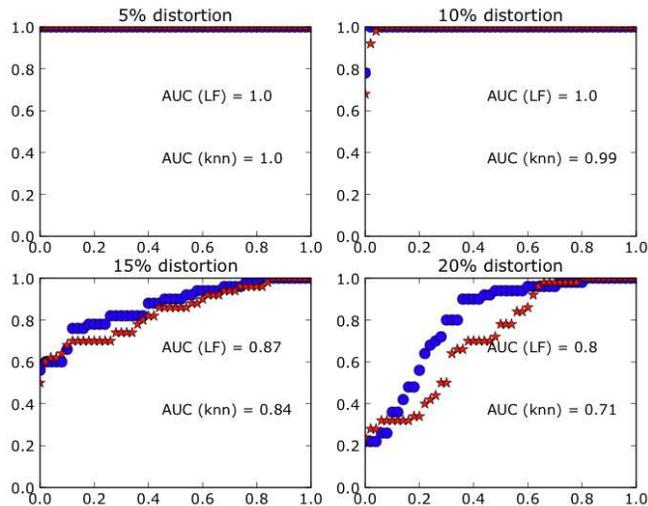}
  \caption{Classifier ROC plots for different distortion levels}
  \label{roccurves}
\end{figure}

The classifiers are evaluated on the basis of the Area under the ROC curve ($AUC$) \cite{fa06}, Blue for $RAG+LF$ and Red for $kNN$ (Figure~\ref{roccurves}). The classification rates of $kNN$ compared with the proposed algorithm is shown in Table 1. As is seen, for low values of distortion, $RAG + LF$ family of classifiers give near ideal performance. For higher noise levels, the algorithm does achieve higher robustness to noise compared to $kNN$.

\begin{center}
\begin{table}[!ht]
\centering 
\begin{tabular}{ | l | l | l | l | l | }
\hline
Distortion \hspace{0.5in} & $5$ \hspace{0.5in} & $10$ \hspace{0.5in} & $15$ \hspace{0.5in} & $20$ \\ \hline
RAG + LF \hspace{0.5in} & 97 \hspace{0.5in} & 97 \hspace{0.5in} & 81 \hspace{0.5in} & 74 \\ \hline
kNN      \hspace{0.5in} & 95 \hspace{0.5in} & 84 \hspace{0.5in} & 72 \hspace{0.5in} & 56 \\ \hline
\end{tabular}
\caption{Classification rates ($\%$) on the synthetic datasets}
\end{table} 
\end{center}

\subsection{Experiments on IAM graph database repository}

A set of experiments were conducted on two standard datasets from the IAM graph dataset repository\cite{rb08}. A brief description of the dataset is reproduced below.

\begin{table}[!ht]
\centering 
\begin{tabular}{l@{\qquad}ccc@{\qquad}ccc}
\hline
\hline
dataset & train,val,test& Classes & max $V$ & max $E$\\
\hline
Letter (HIGH) & 750, 750, 750 & 15 & 8 & 3.1 \\
Fingerprint & 500, 300, 2000 & 4 & 26 & 4.42 \\
\hline
\hline
\end{tabular}
\vspace{1ex}
\caption{Summary of main characteristics of the data sets}
\label{tab:characteristics}
\end{table}

In order to examine the performance of the proposed approach on a two class problem consisting of patterns from morphologically distinct classes, a reduced dataset called \emph{Fingerprint (AW)} was created consisting of patterns belonging to only classes \emph{arch} and \emph{whorl}.



\subsection{Results and discussion}

The state-of-the-art techniques chosen are k-NN (chosen as Reference system)\cite{rb08}, embedding based on Similarity Kernels (SK + SVM), embedding based on Lipschitz Embedding (LE+SVM)\cite{rb09}, and Structurally-described random graphs (SDRG)\cite{ss09}. The approach proposed here is referred to as Random Attributed Graph model + Likelihood as Feature (\textbf{RAG + LF}). RAG + ML denotes the method where a graph pattern is assigned to the class of random prototype graph, which has the maximum likelihood of having generated it. SK+SVM and LE+SVM refer to a family of related classifiers out of which the best performing model is chosen. hence, the comparision is biased towards the same. 

\begin{center}
\begin{table}[!ht]
\centering 
	\begin{tabular}{ | l | l | l | l | }
	\hline
	Method & Letter (HIGH) & Fingerprint (AW) & Fingerprint \\ \hline
	kNN  & 82 & 91.8 & 76.6 \\ \hline

	SK + SVM  & 79.1 & - & 41 \\ 
	LE + SVM  & 92.5 & - & 82.8 \\ \hline

	SDRG  & 64.3 & - & - \\ \hline

	RAG + ML  & 67.2 & 87.5 & 61.1 \\
	RAG + LF  & 75.7$^{\star,\dagger}$ & 95.9 & 78.2$^{\star}$ \\ \hline

	\end{tabular}
\caption{Classification rates ($\%$) on IAM Graph Dataset.  $\star$, $\dagger$ indicates statistically significant improvement of RAG + LF over RAG + ML and SDRG at significance level 0.05 respectively}
\end{table}
\end{center} 

The following observations are made- the results compare well for the Fingerprint dataset overall, and for the Letter (HIGH) dataset compares well with SK + SVM and is superior to SDRG; although k-NN yields good results overall, it faces the computationally challenging task of choosing k. For SK + SVM and LE + SVM, the task of choosing effective prototype set and calculating the graph-edit distance between the dataset and prototype set is expensive as well and offers no analytical insight. The approach presented here is fast as it involves estimating the parameters of random graph model analytically and needs far less graph matching operations corresponding to generating only one class prototype model. The prototypes also give a good summary of node and edge occurence probabilities in the dataset and probability distributions of their attributes. Embedding the prototypes in the space spanned by likelihood values offers statistically significant improvement with almost no significant loss of speed as there fast packages for SVM's and other classification algorithms. 

\section{Conclusions}

This work builds upon the notion of random graph models with applications in structural pattern recognition with the following contributions- with independence assumptions a random attributed graph is represented as a joint random variable in its node and edge occurences and of their respective attribute values, an analytical method to estimate the different probability distributions of a random graph model as a prototype given an ensemble of attributed graphs is presented using a maximum likelihood procedure, the utility of the random graph as a prototype is shown by using the likelihood of an outcome graph as a feature for classification. The proposed approach is suited to contexts involving large number of graph data samples, as determination of random prototype graph is a density estimation problem. It is robust to noise and faster on account of lesser number of graph matching operations that need to be performed in contrast to other approaches.

There are several possible extensions to this approach- first, a method to derive a class of probabilistic clustering and classification algorithms is being currently investigated. This means that the random prototype graph is learned from the dataset in a procedure akin to a standard quantization type scheme. Second, is there a way to tie the classifiers in the feature space directly with the learning of prototypes? To elaborate, it is important to investigate the link between type/family of classifiers on the feature space (due to likelihood) with how the random prototypes are estimated/learned. This would help to integrate probabilistic learning in the domain of graphs with discriminative methods for classification in the subsequent likelihood space. Lastly, the foundations of the random graph definitions needs to be explored- although node and edge independence is useful in that it allows an easy analytical estimation of model parameters, it is too strong an assumption. Is there a way to model dependencies of nodes and edges and their attributes (node/edge co-occurences)? Such a model would help enormously in probabilistic sub-structure analysis methods and also give possibly superior classification and clustering algorithms.


\begin{thebibliography}{[MT1]}

\bibitem[1]{w85}
Wong, A.K.C., You, M.:
Entropy and distance of random graphs with application to structural pattern recognition.
IEEE Trans. PAMI.,Vol-7, No. 5, (1985) 599-609

\bibitem[2]{w80}
Wong, A.K.C., Ghahraman, D.E.:
Random graphs: Structural-contextual dichotomy.
IEEE Trans. PAMI.,Vol-2, No. 4, (1980) 341-348


\bibitem[3]{ss09}
Sole-Ribalta, A., Serratosa, F.:
A structural and semantic probabilistic model for matching and representing a set of graphs.
GbRPR 2009, LNCS 5534, pp. (2009) 164-173



\bibitem[4]{sk93}
Seong, D.S., Kim, H.S., Park, K.H.:
Incremental clustering of attributed graphs.
IEEE Trans. Sys., Man, Cyb., Vol-23, No.5 (1993) 1399-1410


\bibitem[5]{sb95}
Sengupta, K., Boyer, K.L.:
Organizing large structural modelbases.
IEEE Trans. PAMI., Vol-17, No. 4, (1995), 321-332


\bibitem[6]{am98}
Amari, S.-I.:
Natural gradient works efficiently in learning.
Neural Comput., Vol.10, (1998), 251-276

\bibitem[7]{honk08}
Honkela, A., Tornio, M., Raiko, T., Karhunen, J.:
Natural Conjugate Gradient in Variational Inference.
Neural Information Processing, (2008), Springer, 305-314

\bibitem[8]{fr08}
Fischer, A., Riesen, K., Bunke, H.:
An experimental study of graph classification using prototype selection.
ICPR, IEEE, (2008), 1-4


\bibitem[9]{ht99}
Hollmen, J., Tresp, V., Simula, O.:
A self-organizing map algorithm for clustering probabilistic models.
ICANN'99, IEE, Vol. 2, (1999), 946-951


\bibitem[10]{dh00}
Duda, R.O., Hart, P.E., Stork, D.G.:
Pattern classification, Wiley-Interscience, (2000)


\bibitem[11]{hauss98}
Jaakkola, T., Haussler, D.:
Exploiting generative Models in discriminative classifiers.
Advances in Neural Information Processing Systems, (1998), 487--493

\bibitem[12]{gr96}
Gold, S., Rangarajan, A.:
A graduated assignment algorithm for graph matching.
IEEE Trans. PAMI, Vol. 18, No. 4, (1996), 377-388

\bibitem[13]{bh08}
Ben-Hur, A.:
PyML- A Python Machine Learning package, (2008)


\bibitem[14]{fa06}
Fawcett, T.:
An introduction to ROC analysis.
Pattern Recogn. Lett., Vol. 27, No.8, (2008), 861-874


\bibitem[15]{rb08}
Riesen, K., Bunke, H.
IAM Graph database repository for graph based pattern recognition and machine learning.
SSPR+SPR '08, Springer, (2008), 287-297


\bibitem[16]{rb09}
Riesen, K., Bunke, H.
Graph classification by means of Lipschitz embedding.
IEEE Trans. Sys. Man Cyber. Part B, Vol. 39, No. 6, (2009), 1472-1483







\end{thebibliography}
\end{document}